\documentclass[11pt,a4paper]{article}
\usepackage{times,latexsym}
\usepackage{url}
\usepackage[T1]{fontenc}

\usepackage{graphicx}
\usepackage{booktabs, multirow}
\usepackage{multirow}
\usepackage{enumitem}
\usepackage[at]{easylist}
\usepackage{amssymb}
\usepackage{textcomp}

\usepackage[acceptedWithA]{tacl2021v1}

\usepackage{xspace,mfirstuc,tabulary}

\newif\iftaclinstructions
\taclinstructionsfalse 
\iftaclinstructions
\renewcommand{\confidential}{}
\renewcommand{\anonsubtext}{(No author info supplied here, for consistency with
TACL-submission anonymization requirements)}
\newcommand{\instr}
\fi

\iftaclpubformat 

\else

\fi

\title{NLP Security and Ethics, in the Wild} 

\author{Heather Lent$^{1}$ \hfill Erick Galinkin$^{2}$ \hfill Yiyi Chen$^{1}$  \\    
\textbf{Jens Myrup Pedersen$^1$ \hfill Leon Derczynski$^{2,3}$ \hfill Johannes Bjerva$^{1}$} \\
          $^{1}$Aalborg University, Denmark \\
          $^2$NVIDIA Corporation, 
          $^3$IT University of Copenhagen, Denmark \\
           \texttt{\{hcle, jbjerva\}@cs.aau.dk} \\
}

\date{}

\begin{document}
\maketitle
\begin{abstract}
As NLP models are used by a growing number of end-users, an area of increasing importance is NLP Security (NLPSec): assessing the vulnerability of models to malicious attacks and developing comprehensive countermeasures against them. While work at the intersection of NLP and cybersecurity has the potential to create safer NLP for all, accidental oversights can result in tangible harm (\textit{e.g.}, breaches of privacy or proliferation of malicious models). In this emerging field, however, the research ethics of NLP have not yet faced many of the long-standing conundrums pertinent to cybersecurity, until now. We thus examine contemporary works across NLPSec, and explore their engagement with cybersecurity's ethical norms. We identify trends across the literature, ultimately finding alarming gaps on topics like harm minimization and responsible disclosure. To alleviate these concerns, we provide concrete recommendations to help NLP researchers navigate this space more ethically, bridging the gap between traditional cybersecurity and NLP ethics, which we frame as ``white hat NLP''. The goal of this work is to help cultivate an intentional culture of ethical research for those working in NLP Security.
\end{abstract}

\section{Introduction}
Securing large language models (LLMs) and NLP technology in general has not been a priority until recently.
Yet mass adoption of this technology has led to deployment in contexts where security failures present a risk to individuals, organizations, and society at large (demonstrated by \textit{inter alia} LLM-assisted identity fraud \cite{ackerman-news-22}, phishing campaigns \cite{hazell2023spear}, and automated influence operations \cite{goldstein2023generative}).
If the latest NLP technologies are to withstand an increasing barrage of threats, NLP practitioners must now educate themselves on cybersecurity and cybersecurity practitioners must educate themselves on NLP.
Yet in this process, as one culture of research adapts to another, there is potential for long-standing intra-community norms to be lost in translation, including ethical norms. 
In interdisciplinary research, an accidental oversight of ethical norms from one field can risk \textit{reintroducing} previously resolved ethical dilemmas.

Discussions around ethical research conduct have long been a concern for both 
cybersecurity~\cite{molander1998legitimization, himma2008handbook, matwyshyn2010ethics, menloreport, christen2020ethics, Kohno-2023}
and for NLP~\cite{wiener1960some, samuel1960some, dennett1997hal, 1667948, anderson2007machine, hovy-spruit-2016-social, leidner-plachouras-2017-ethical}, but given that these disciplines have historically been disjoint fields, the specific ethical and sociocultural norms of both fields have developed in separate silos. 
To better understand how interdisciplinary NLPSec has adapted to the ethical norms and values across both disciplines,
we examine a set of peer-reviewed NLPSec publications from NLP venues to gauge the compliance with norms in cybersecurity.
We find that several principles regarded by the cybersecurity community as best practices have not been widely adopted in NLPSec research, despite measures in the NLP peer-review process to improve research practices. 
Simultaneously, we find that NLP ethical norms regarding lower-resourced languages are at risk of being overlooked in NLPSec. 
To help save NLPSec the potential growing pains of reinventing the wheel, in this work, we seek to address this issue through an interdisciplinary conversation about ethics, with the goal of cultivating a culture of \emph{white hat NLP}.

\begin{figure}
    \centering
    \includegraphics[width=\linewidth]{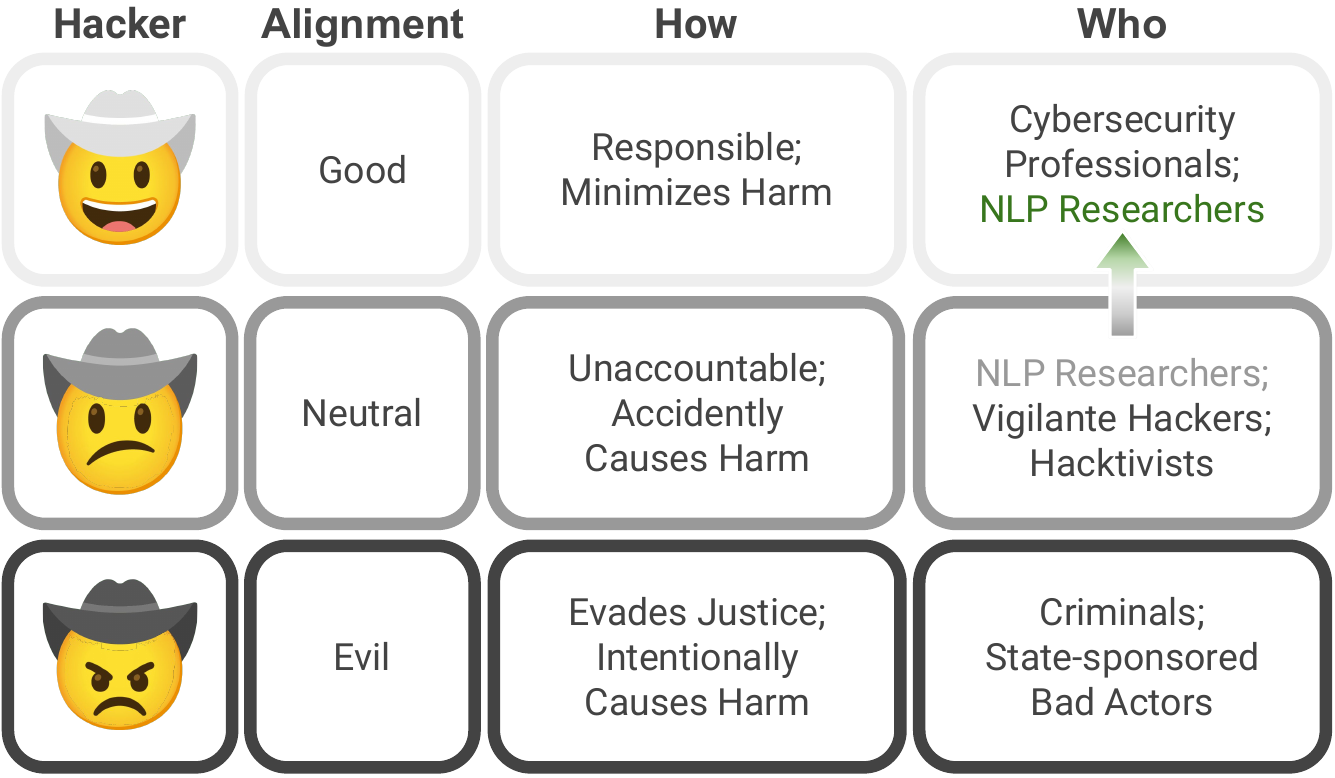}
    \caption{We argue current works across NLPSec occupy a gray area in research ethics, comparable to the cybersecurity concept of a ``gray hat'' hacker. In this work, we provide concrete recommendations to help move the field of NLPSec towards more ethically-grounded research practices.
    }
    \label{fig:firstfig}
\end{figure}

\paragraph{Ethical NLP Security}
In cybersecurity, a ``white hat'' hacker is typically a professional hired by a company with the specific purpose of maintaining or increasing the existing security of a computer system. 
White hat hackers\footnote{This verbiage has long been common parlance in cybersecurity, though it has recently been criticized for its connoted colorism. As there is no widely adopted alternative, we trust readers to accept our good faith usage of the term.}, referred to more generally as ``ethical hackers'', are keen to engage with vendors -- either their employer or a third party -- whose products they have found flaws in. 
They aim to get security issues fixed quickly and release details of their findings to the public so defenders can evaluate their own environments and prioritize concomitant patching and mitigation efforts.
In contrast, a ``black hat'' hacker represents a cybercriminal, and somewhere in between is the ``gray hat'' hacker, who infiltrates others' computer systems without permission, with the intention of enhancing security~\cite{Falk2004GRAYHH}. 
This is generally mapped to similar activity in the context of LLMs, where  LLM red teaming is defined as a limit-seeking, manual, and non-malicious activity~\cite{inie2025summon}.
Outside the strict boundaries of ethical hackers, gray hat hackers rely on their own moral compass, which can lead to potentially precarious situations, like introducing the risk of authorized system intrusions~\cite{christen2020ethics}. 
Alarmingly, much of the contemporary work in NLPSec is arguably similar to the above gray hat scenario (Figure~\ref{fig:firstfig}). Due to the distributed and decentralized nature of the research landscape, most NLPSec researchers will necessarily lack a mandate from organizations to investigate security vulnerabilities inherent to models. Researchers also lack direct accountability to those most affected by the security breaches they study. 
While the ACM Code of Ethics\footnote{\url{https://www.acm.org/code-of-ethics}} provides overarching guidance, it naturally cannot provide exact guidance for every ethical conundrum, leaving individual researchers to rely on their own moral compass, like a gray hat hacker. 
It is in this light that we aim to extend the framework of ``white hat'' to NLP. We thus define the scope of white hat NLPSec to consist of \textit{works which are intentionally and carefully grounded in the established ethical best practices of both cybersecurity and NLP}.

\paragraph{Contributions}
We examine the current culture of research ethics in NLPSec through a survey of 80 peer-reviewed works across NLP venues, measuring their compliance with typical ethical practices from cybersecurity. 
To start, we introduce these ethical practices, and touch upon their relevance to works in NLPSec (Section~\ref{sec:cybersec}). 
We then describe our selection process for papers included in the survey, and our annotation process identifying compliance with these cybersecurity ethical best practices (Section~\ref{sec:method}).
Concretely, we find that ethical norms like harm minimization and responsible disclosure are not widely adopted in NLPSec (Section~\ref{sec:nlp}).
We identify several challenges of adopting ethical norms from both cybersecurity \textit{and} NLP to the field of NLPSec, and we
provide concrete advice to help researchers engage in research \textit{more ethically} than previously (Section~\ref{sec:discussion}).
While the recommendations presented in this work cannot provide solutions to all potential ethical conundrums latent to NLPSec, the discussions in this work 
aim to highlight the urgency for discussion on research ethics in NLPSec and serve as a conversation starter.

\section{The Culture of Ethics in Cybersecurity}\label{sec:cybersec}

Within any field, some research may have negative consequences, and foreseeable risks generally serve as a guiding force for calibrating ethical norms. 
This holds particularly true for cybersecurity, as a field that is largely engaged with preempting and outmaneuvering criminals while protecting the public. 
As the field is naturally situated in a sensitive context, cybersecurity's research culture has unsurprisingly evolved to be one that often faces difficult ethical discussions \cite{menloreport}. 
In this section, we will introduce three ethical norms commonly expected of works across cybersecurity. 
For each norm, we further describe potential nuances, borrowing relevant trolley problems directly from \citet{Kohno-2023}.
Our aim is to familiarize readers with these concepts, demonstrate how best practices in cybersecurity continue to be developed, and briefly touch upon how these concepts are relevant to works in NLPSec. 

\subsection{Harm Minimization}
A familiar concept to cybersecurity practitioners is that the ``maliciousness'' of tools is context-dependent.
It is well-documented that black hat hackers often misuse legitimate software~\cite{10.1145/3043955, menloreport} for their own purposes.
For instance, remote desktop management software is commonly used to facilitate remote troubleshooting for users.
However, these very same tools can be used by malicious actors to maintain a foothold in a victim network.
A critical part of cybersecurity methodology is establishing normal, secure usage patterns for these tools to minimize the potential risk associated with them.

To minimize harm, one must first identify the potential dangers.
For example, \citet{10199005} aim to examine the impact of the ongoing Russia-Ukraine war on Ukrainian critical infrastructure by scanning web traffic.
They observe that one potential risk inherent to such work includes the accidental disruption of end systems, which could exacerbate the current plight of non-combatant Ukrainians under the war. 
To minimize harm, ~\citet{10199005}'s methodology ensures uninterrupted web traffic, allows end-users to opt-out of scans, and safeguards sensitive data recovered by the scan, meanwhile collaborating with an NGO to help with responsible disclosure to Ukrainian authorities before publication. 
Here, potential victims of cyberattacks are protected to the greatest extent possible, largely through strategic choices laid out in the methodology. 

While causing some amount of harm -- however inadvertent -- may be inevitable, grappling with these potential ethical considerations early on can help researchers to avoid ethical dilemmas and minimize that harm.
To this end, \citet{Kohno-2023} propose a security trolley problem: 
in the context of an AI-based employment software tool, if a data breach compromised sensitive user data, should security researchers study the leaked data (prioritizing a potential benefit to the public, if unfair bias can be established from the data) or not study it (prioritizing the affected users' right to privacy)? 
\citet{Kohno-2023} demonstrate that both conclusions can be justified through moral philosophy frameworks, specifically consequentialist and deontological ethics\footnote{In~\citet{Kohno-2023}, utilitarianism is adopted to conduct consequentialist analyses, centering the outcome of an action, whether it produces the greatest net positive well-being; the deontological analyses follow Kantian moral philosophy, whereby humans, as rational beings, have an absolute moral duty to justice regardless of consequences.}, respectively. The goal of this exercise is not to encourage moral relativism; on the contrary, it is to stress the necessity for continuous and principled conversations on research ethics in cybersecurity.
This discussion reinforces the notion that much research in cybersecurity will demand researchers to assess not only the potential harm inflicted by their work, but also the potential harm inflicted by foregoing a study. 
In the absence of strict governing boards, who grant permission to researchers for specific studies, it remains the duty of an ethical researcher to remain vigilant to this characteristic of cybersecurity research and adapt accordingly. Other common strategies for harm minimization include anonymizing published datasets \cite{mirsky2016sherlock}, limiting what information is published \cite{burstein2008conducting}, and foregoing a study entirely \cite{macnish2020ethics}. Ultimately, these strategies reflect a culture of research ethics that has been developed over time.

Similar to cybersecurity, the list of potential harms considered in NLPSec can range from immediate misfortunes (\textit{e.g.}, models inverted to reveal private data or prompt injection leading to remote code execution) to systemic harms associated with AI systems (\textit{e.g.}, LLMs being weaponized to generate hate speech about a marginalized group). 
We explore in detail the potential risks identified by NLPSec researchers in Section~\ref{subsec:harms} and the implications of harm minimization in NLPSec in Section~\ref{subsec:multilinguality}.

\subsection{Coordinated Vulnerability Disclosure}\label{sec:cvd}
In traditional cybersecurity, a key aspect of ``white hat'' ethical hacking is the clear and timely disclosure of relevant information through a process known as coordinated vulnerability disclosure (CVD)~\cite{ISO29147}.
In CVD, when an issue is found, it is reported to the vendor on a best-effort basis \textit{before} findings are published.
This gives those responsible for the affected software an opportunity to address the discovered issue ahead of any public disclosure, often on some agreed-upon timeline.
When the vulnerability is disclosed by a vendor, researcher, or in a joint release, information about remediation or prevention is included so affected parties can minimize or mitigate their exposure to the issue. 
To this end, CVD offers security researchers a way to maintain transparency, without assisting malicious actors, as the published vulnerabilities will hopefully have already been patched. 

In the real world, of course, CVD can be complicated by a variety of factors. 
For example, \citet{Kohno-2023} detail another pertinent trolley problem: imagine an industrial researcher is assigned to review an anonymous manuscript, which reveals a severe security vulnerability in software supported by the industrial researcher's employer. 
Should the researcher disclose the security vulnerability to their employer (thus prioritizing the safety of a large number of end-users, who would otherwise be exposed) or not disclose it (thus prioritizing the authors' rights with respect to peer-review)?
As pointed out by the Menlo Report \cite{menloreport}, in such circumstances, different stakeholders are likely to have different priorities, and sometimes the most ethical action may be in direct conflict with one's best interests. Accordingly, the above trolley problem can be reexamined from the perspective of different stakeholders to even further explore the ethical consequences of one decision over another. 
Perhaps this ethical conundrum could have been avoided entirely, if the imagined authors had planned to do CVD from the start.
In most cases, however, cybersecurity practitioners are highly incentivized to complete CVD. From bug bounty programs to vulnerability disclosure leaderboards, CVD is a well-established norm, which helps companies and organizations secure these systems, while being highly prestigious for researchers.

In the context of NLPSec, at face value, CVD is relevant to works examining security vulnerabilities of proprietary language technologies.
We discuss the presence and ramifications of CVD in NLPSec in Sections~\ref{subsec:nocvd} and ~\ref{sec:cvd-revisited}, respectively. 

\subsection{Public Disclosure}
Another common practice in cybersecurity as part of, or -- less desirably -- in lieu of, CVD is public disclosure.
The essential idea is that attackers, particularly those motivated by criminal or national security desires, are incentivized to take a potentially interesting or useful vulnerability and uncover ways to exploit it.
In contrast, defenders are generally already busy triaging alerts, responding to incidents, managing defensive infrastructure, and other important tasks that constrain their ability to develop a way to test for and detect potentially vulnerable systems.
In other words, defenders, stymied by other responsibilities, may move at a much a slower pace than attackers.
As reported by \citet{condon2022vulnerability}, more than half of widely exploited vulnerabilities were leveraged by attackers against victims in less than one day after disclosure.
This puts defenders at a distinct disadvantage in terms of the speed at which they can react to emerging threats, placing the fate of their security in the hands of software vendors and leaving them at the mercy of their own scheduled patching cycles. 
In this way, defenders can be informed of potential threats quickly and benefit greatly from the help of white hat hackers who share their findings in a responsible manner.

One such example where public disclosure may be called for, in lieu of CVD, is in the case a vulnerability is identified in relation to a company that no longer exits and so there is no entity with which to coordinate. 
Such a scenario is not unthinkable in the real world, as \citet{Kohno-2023} introduce another security trolley problem whereby security researchers have identified a vulnerability in an imagined medical device that is embedded in sizable population of patients but is no longer serviceable, as the manufacturer is out of business.
Should the researchers publicly disclose the vulnerability (thus prioritizing the patients' right to informed consent and bodily autonomy)
or not disclose it (prioritizing the patients' peace-of-mind and happiness)? 
Again, \citet{Kohno-2023} demonstrate that both decisions can be reached \textit{via} different frameworks of moral philosophy; the purpose of this illustration is equivocally \textit{not} to show that any decision taken by the researchers can be simply justified after the fact, but rather to again underline the importance of a cybersecurity research landscape, which is actively engaged in conversations on ethics. 
In practice, similar scenarios to the above trolley problem have led to the formation of Computer Emergency Response Teams (CERTs). If a practitioner identifies a bug which has no obvious path for CVD, and the bug is likely to be abused by malicious actors upon public disclosure, they may opt to disclose the bug only to a CERT or to a trusted community. The creation of such CERTS or trusted communities in critical spaces again underscores that the act of disclosure is highly important and carefully considered across cybersecurity.

In NLPSec, publication in open-access venues such as the ACL or ArXiv is a form of public disclosure. 
Like defenders in cybersecurity, the majority of NLPSec researchers have limited resources, putting them at a relative disadvantage to bad actors. In this way, researchers can help each other by open-sourcing code, where appropriate.
We examine the prevalence of fully open-source works in Section~\ref{subsec:closedsource} and revisit public disclosure in Section~\ref{subsec:dualuse}.

\section{Methodology}\label{sec:method}
We examine 80 pertinent, peer-reviewed works across NLPSec for common trends and themes pertaining to discussions on research ethics. 
All papers were manually gathered from the ACL Anthology\footnote{\url{https://aclanthology.org/}}, by querying keywords associated with common attack types (\textit{i.e.}, ``security'' + $\{$``adversarial'', ``backdoor'', ``data recreation'', ``inversion'', ``instance encoding``$\}$) to ensure they fall within the scope of NLPSec. 
 We specifically did not include ``attack'' or ``defense'' in our keyword search, to avoid influencing the results.
For each keyword search, we examined the relevance-sorted list of results. 
As terms like ``adversarial'' or ``inversion'' can be used in a wide variety of contexts beyond NLPSec, we first review each paper title, and keep those which are obviously relevant to NLPSec; where relevance is unclear from the title alone, we further scan the abstract to determine relevance. 
Papers dated between 2019--2023 were obtained from the anthology in January 2024 (n=60).
To include papers published in 2024, we used the same procedure in November 2024, following the publication of the EMNLP 2024 proceedings (n=20). 
Accordingly, all 80 papers for this study are guaranteed to be relevant and peer-reviewed. 
While a substantial number of publications in NLPSec -- particularly those concerning attack methods -- are published in preprint archives like ArXiv, such papers do not necessarily go through peer review (e.g., \citet{zou2023universal,mehrotra2023tree}).
We intentionally limit ourselves to peer-reviewed publications to ensure rigorous publications whose claims have been vetted, rather than considering preprints whose claims we must take at face value.
Additionally, publications in major conferences and journals are obliged to abide by the ACM Code of Ethics; as such, we assume good intent by the authors to meet a standard of ethics that is acceptable to the scientific community.
For each of the 80 papers, we manually annotate the following:

\paragraph{Attack Scenario.} (Values: \texttt{Adversarial}, \texttt{Backdoor}, or \texttt{Data Reconstruction attack}). This is coded in accordance with the keyword which was used to retrieve the paper. 

\paragraph{Main contribution.} (Values: \texttt{Attack}, \texttt{Defense}, or \texttt{Both}). We assign this value based on the text of the title and abstract only. For example, a paper which discusses exclusively an attack method in the title and abstract is coded as \texttt{Attack}, even if a defense is offered later in the paper, for example in the list of contributions, or in an analysis. 
In this way, our coding process intends to mirror the authors' framing of their own work.  

\paragraph{Discussion of Ethical Concerns.} (Values: \texttt{Yes}, \texttt{No}). Here, we first look for the presence of a dedicated ethics section; if the paper has one, it is coded \texttt{Yes}. If the paper does not have such a dedicated section, we continue to search for discussions on ethics by looking for a broader impacts section, then reading the conclusion, the limitations, and the introduction, as these sections typically contain high-level reflections on topics such as ethics. If a discussion of ethics has still not been identified in the aforementioned sections, we finally search the document for the lemma ``ethic'' and examine all possible matches to determine whether the paper discusses ethics. Otherwise, if there are no matches, the paper is coded as \texttt{No}.

\paragraph{Dual Use and Misuse.} (Values: \texttt{Yes}, \texttt{No}). The coding process for dual use and misuse occurs in step with that of the ethics discussion. 
That is, we first check the ethics section (if it exists), as this is where an outright dialogue on misuse is most likely to occur. 
If we do not find it there, we then check the conclusion, limitations, and introduction, accordingly. 
If we still have not located discourse on dual use or misuse in these sections, we search the full document for the lemmas ``use'', ``leverage'', and ``malicious'', and check the context of any matches. If no discussion has been identified through this process, the paper is coded \texttt{No}. 
Note that we annotate dual use as misuse separately (definitions in Section~\ref{subsec:dualuse}). 

\paragraph{Coordinated Vulnerability Disclosure.} (Values: \texttt{Yes}, \texttt{No}). For this variable, we check the introduction, conclusion, ethics/broader impact, limitations, and also footnotes. 
If CVD is not identified, we search for the following lemmas: ``disclose'', ``contact'', ``reach'', ``communicate'', and ``company'', in an attempt to locate discussions on coordinated vulnerability disclosure. If still nothing is found, the paper is coded \texttt{No} for CVD. 

\paragraph{Open-Source Code.} (Values: \texttt{Yes}, \texttt{Empty}, \texttt{No}). Open-source papers typically link to the project's Github page in a footnote on the first page. 
Accordingly, we search for ``github'', and cross-reference any repositories with the associated footnote, to confirm that the linked code is contributed by the authors. 
When a Github repository is found, we follow this link to examine the availability and contents of the repository. 
If the link is broken, we code the value \texttt{Empty (Broken Link)}. If the repository contains only a \textsc{README} and no scripts, we code it as \texttt{Empty (Empty Repository)}. 
When a repository cannot be located through the footnotes, we further search the following lemmas: ``code'', ``provide'', and ``publish'', and check the surrounding context, before ultimately coding the paper \texttt{No}, for no code.    

\paragraph{Other Metadata.} This includes author \textbf{affiliations}, \textbf{datasets} used, \textbf{languages} involved, and the \textbf{models} attacked. For each paper, we document the unique set of affiliations and their associated countries. Datasets are collected from the descriptions of experiments and corroborated directly against tables presented in a given work. Accordingly, we also record the languages present for each dataset, as described by the paper in the experiments, or otherwise inferred from the scope of the paper; most papers work solely on English alone, and those with a wider scope of languages tend to very clearly document this. Victim models are identified in the same manner as datasets, through the description of the experiments and the associated tables.

\begin{figure}
    \centering
    \includegraphics[width = \linewidth]{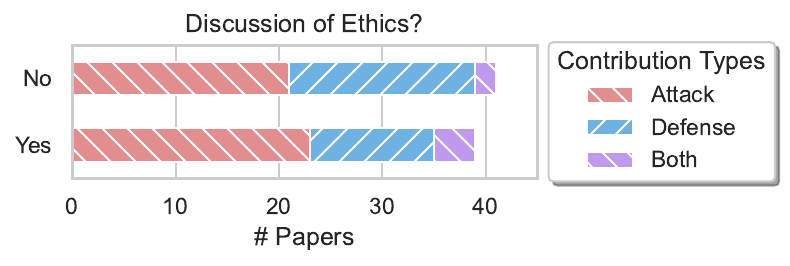}
    \caption{Only half of sampled works across NLPSec include no discussion of ethics (41/80), even when introducing attacks that could be misused by bad actors, demonstrating the pressing need to assess and discuss ethics in this space.}
    \label{fig:ethicsXcontribution}
\end{figure}

Resulting from the annotation process, in Figure~\ref{fig:ethicsXcontribution}, we observe that approximately half (51\%) of the works include no discussion on ethical considerations, 
underscoring the critical need for a broader dialogue on research ethics in NLPSec. 
Note that these findings refer to our sample, covering papers vetted by reviewers which should adhere to established ethical standards. 
We fear that the situation 
beyond reputable, peer-reviewed venues may be substantially worse.
For the full list of sampled papers and their associated annotations, we refer readers to Appendix~\ref{appendix:papers} Tables~\ref{tab:adversarial}, ~\ref{tab:backdoors}, and ~\ref{tab:data reconstruction}. 
Additionally, Appendix~\ref{appendix:details} provides notable metadata concerning the sampled papers, such as years and venue of publication (Figures~\ref{fig:year} and \ref{fig:venue}), the global distribution of author affiliations (Figure~\ref{fig:authors}), and the most used datasets (Figure ~\ref{fig:datasets}).  

\subsection{Survey Coverage}
\begin{figure}[t]
    \centering
    \includegraphics[width = \linewidth]{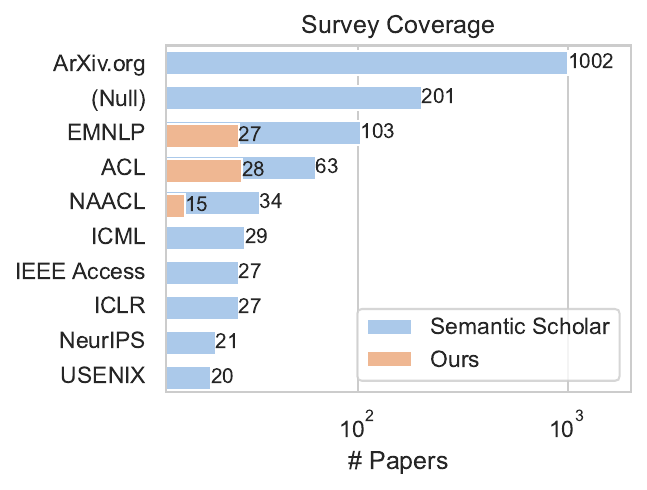}
    \caption{An approximation of the broader field of NLPSec, across the 10 most frequent venues.
    Where multiple venues exist for a single paper, the Semantic Scholar API prioritizes publisher venues (\textit{e.g.}, ACL) over preprint repositories (\textit{e.g.}, ArXiv), when available.
    A ``Null'' value is returned by the API when Semantic Scholar lacks reliable venue metadata for that paper.
    Against this approximation, we also show a subset of our 80 sampled papers, which specifically belong to the displayed venues (``Ours''). Here, we include Findings papers in the counts for EMNLP, ACL, and NAACL, as the API did not distinguish between them.}
    \label{fig:semanticscholar}
\end{figure}

To ensure that our sample of 80 papers is representative of works across NLPSec, we employ a citation crawler\footnote{\href{https://gist.github.com/hclent/682e87c4a8f72f9af95b84c7438a32bf}{https://gist.github.com/hclent/}} through the Semantic Scholar API \cite{Kinney2023TheSS}. We begin with 11 seed papers, widely cited across NLP \textit{(i.e.} \citet{Mikolov2013EfficientEO, Mikolov2013DistributedRO, Devlin2019BERTPO, Sanh2019DistilBERTAD, Liu2019RoBERTaAR, Raffel2019ExploringTL, workshop2023bloom176bparameteropenaccessmultilingual, touvron2023llamaopenefficientfoundation, touvron2023llama2openfoundation, jiang2023mistral7b, Achiam2023GPT4TR}). 
The citation networks for the seed papers points us to 223,078 citing works (as of December 2024), which can be considered to be broadly topical to NLP. 
From the resulting citations, we apply additional filters, in order to further narrow the scope from NLP to NLPSec. Concretely, we
check each paper's title and abstract for matches with the following lemma: ``secur'', ``attack'', ``defen''. 
Again, we intentionally avoid ambiguous search terms like ``adversarial'', as they are widely used outside of an NLPSec context. 
Through this filtering process, we identify 2,782 unique publications, which gives us a coarse-grained estimation of the broader scope of NLPSec. 
These citations are dated from 2020-2024, with a reasonable share at ACL* venues. 
We find that 70 of our sample papers are present in the $\sim$200 works at ACL* venues  (Figure~\ref{fig:semanticscholar}), demonstrating that our survey represents upwards of 35\% of works in main ACL* venues, providing a healthy sample size to draw conclusions on trends pertaining to ethics across NLPSec. 

\section{NLPSec Survey Results}\label{sec:nlp}

Despite NLPSec's position as an interdisciplinary field, we find that the works in our survey have largely failed to adopt ethical norms of cybsersecurity research. 

\subsection{NLPSec Disagrees on Potential Harms}\label{subsec:harms}

\begin{figure}[t]
    \centering
    \includegraphics[width=\linewidth]{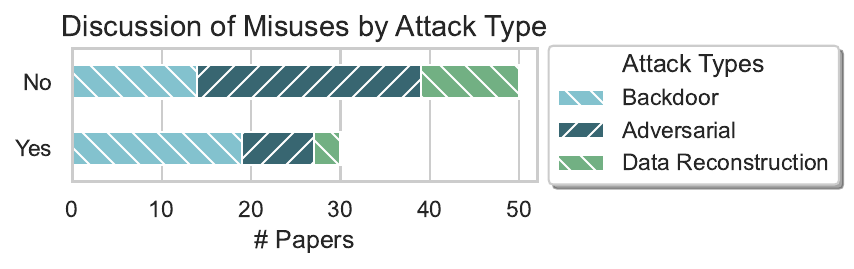}
    \caption{Across our sample, most papers do not mention or discuss the potential for misuse. No works discuss dual use.}
    \label{fig:misuse}
\end{figure}

The attempt to minimize harm requires first an assessment of what constitutes harm, in a given context.
We find that explorations of risk assessment vary widely across surveyed NLPSec works. 
Many cite \textit{misuse} -- the potential for malicious actors to weaponize their proposed methods or code towards criminal ends -- as harm that could possibly arise as a result of public disclosure (\textit{e.g.},~\citet{xu-etal-2021-grey, zeng-etal-2021-openattack, li-etal-2023-multi-target}).
Though, the recognition of misuse as a potential harm varies across attack types within our sample, with works related to Backdoor attacks being the most concerned, and works related to Adversarial attacks being the least (see Figure~\ref{fig:misuse}). 
At the same time, some authors assert that there are \textit{no} inherent risks to their work (\textit{e.g.}, \citet{liu-etal-2023-maximum, zhang-etal-2022-dim}). 
Such claims of risk-free research are more often found in manuscripts introducing defense mechanisms, as the authors may mention how defenses are 
unlikely to be misused by the public (\textit{e.g.}, \citet{qi-etal-2021-onion}), impossible to misuse (\textit{e.g.}, \citet{jin-etal-2022-wedef}), or even morally noble (\textit{e.g.}, \citet{li-etal-2021-bfclass-backdoor}). 
In direct contrast,~\citet{yang-etal-2021-rap} discuss the liability that their proposed defense mechanism could be directly studied by bad actors, wishing to sidestep such safety measures. 

The above findings illuminate that \textbf{NLPSec researchers disagree on \textit{what} potential harms may exist, and \textit{whether} potential harms exist at all.} 
How can the norm of harm minimization thus be adopted in NLPSec, in the context of such nonagreement?; and who is at risk of suffering harms, if this nonagreement is left unresolved? 
In Section~\ref{subsec:dualuse}, we expand on misuse  NLPSec in more detail; in Section~\ref{subsec:multilinguality}, we discuss how current trends in NLPSec research further jeapordize vulnerable communities.

\subsection{NLPSec Lacks a Culture of CVD}\label{subsec:nocvd}
\begin{figure}
    \centering
    \includegraphics[width=\linewidth]{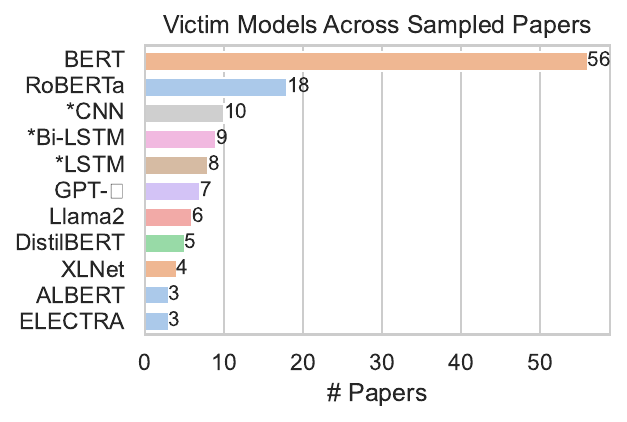}
    \caption{Distribution of victim (or suspect) models, across the sampled works in NLPSec. Models prepended with an asterisk ``*'' are ones trained individually, rather than downloaded from pre-trained weights. For the purposes of this annotation, we do not disambiguate between model sizes, for example, BERT-large versus BERT-small. 
    Similarly the GPT-$\square$ label represents the set of all GPT-based models across our sample (\textit{i.e.}, \texttt{text-embeddings-ada-002}, GPT-2, GPT-J, GPT2-XL, GPT3.5, and GPT-NEO1.3).
    For visualization purposes, we exclude the long tail of models that have been attacked (or defended) by only one or two papers.
    }
    \label{fig:victim}
\end{figure}

Among our sample of 80 works in NLPSec, we find no outright declarations of CVD.
In part, this can likely be explained by the kinds of victim models researchers choose to experiment on (see Figure~\ref{fig:victim}). Some are simple, self-trained models, without pre-trained weights (e.g., Bi-LSTM's), while others are long-time, staple LM's (e.g., BERT~\cite{devlin-etal-2019-bert} and RoBERTa \cite{Liu2019RoBERTaAR}), which no longer represent the state-of-the-art.
Still, as models like BERT and RoBERTa are still in heavy circulation, with official versions in circulation from businesses (Google and Meta, respectively), they are candidates for CVD, at face value.
Moreover, works engaging with newer, proprietary models (\textit{e.g.}, those from OpenAI), still do not state clearly whether or not CVD occurred as a part of the publication process.
Thus it is clear that the ethical norm of CVD in cybersecurity has not yet reached the world of NLPSec, which entails alarming ramifications: \textbf{there is an opportunity for cybercriminals to weaponize the security vulnerabilities revealed by works in NLPSec}.
We examine potential challenges for CVD in NLPSec in Section~\ref{sec:cvd-revisited}). 

\subsection{NLPSec Falls Short on Public Disclosure}\label{subsec:closedsource}
\begin{figure}[t]
    \centering
    \includegraphics[width = 0.8\linewidth]{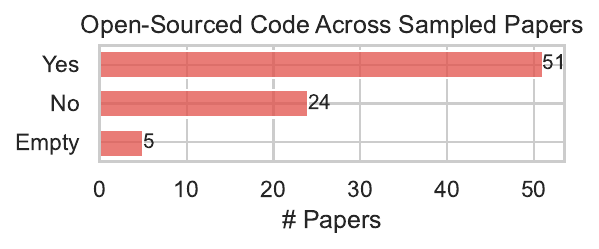}
    \caption{Proportion of papers with open-source repositories from our sample of peer-reviewed papers. 
    }
    \label{fig:opensource}
\end{figure}

While researchers may not agree on the presence or severity of hazards resulting from the publication of their work, it is generally accepted that work in this domain is justified by the need to ``raise awareness'' of newly uncovered security vulnerabilities (\textit{e.g.}, \citet{yang-etal-2021-rethinking, qi-etal-2021-mind, qi-etal-2021-hidden, chen-etal-2022-textual}). 
If we accept the norm from cybersecurity  --
that a full public disclosure should typically come with the necessary code to assist other ethical researchers and practitioners -- NLPSec falls short of this ideal. 
From our sample of works, 36\% are functionally closed-source (see Figure~\ref{fig:opensource}). 
As a consequence, \textbf{white hat NLPSec researchers and practitioners may be at a disadvantage when it comes to securing systems}.
Depending on the severity of a vulnerability discussed in a given work, malicious actors 
may be able to benefit from the latency of white hat engineers re-implementing a paper's threat model.  
The relationship between public disclosure and harm minimization is explored in Section~\ref{subsec:dualuse}.

\section{Discussion}\label{sec:discussion}
While NLPSec is an interdisciplinary field with an indisputable connection to cybersecurity, there are cases where the parallels between these fields diverge.  
In this section, we explore some areas where the analogy between NLPSec and cybersecurity fails, with the goal of illuminating the urgent need for a broader conversation about ethics in NLPSec. 
To help initiate such a conversation, we conclude with some concrete recommendations for NLPSec researchers, 
towards adopting better practices for more ethical NLPSec research.  

\subsection{To Name It Is To Own It: Misuse and Other Harms}\label{subsec:dualuse}
In our survey, \textit{misuse} was the most commonly cited potential harm inherent to research in NLPSec (37\% of works in Figure~\ref{fig:misuse}). The threat of misuse can be understood to result from public disclosure, as malicious actors are known weaponize information therein \cite{rapid-2022-rapid4}. 
With the goal of preventing misuse, one reactionary approach would be to resign ourselves from publicly sharing such sensitive work (\textit{i.e.}, \textit{security through obscurity} ~\cite{guo2018defending}).
This approach has been largely rejected by the wider cybersecurity community, as security through obscurity is difficult to maintain, creates a false sense of security, and clashes with the scientific value of transparency (\textit{e.g.}, \citet{cryptoeprint:courtois2009dark}).
Works in NLPSec are thus caught in what has been dubbed \textit{The Devil's Triangle} \cite{thieltges2016devil, leidner-plachouras-2017-ethical}: the path towards model security hinges upon transparency, which is required for researchers to make progress, but is also advantageous for cybercriminals, while innocent actors can be harmed in the cross-fire.
To help NLPSec escape the Devil's Triangle, we look to existing works in NLP on misuse, and examine whether these suggestions make sense in NLPSec. 

\paragraph{Dual Use and Misuse in NLP}
NLP technologies like LLMs can be leveraged for a wide variety of applications, ranging from the virtuous (\textit{e.g.}, improving accessibility), to the reprehensible (e.g., proliferating hate speech) and the innocuous (\textit{e.g.}, generating fan fiction). 
That these models can simultaneously be utilized for \textit{both} ``legitimate'' and ``illegitimate'' purposes is commonly referred to as \textbf{dual use} \cite{10.1093/ejil/chad039}. 
Traditionally, dual use has been viewed primarily through the lens of a ``civilian'' versus ``military'' dichotomy in terms of applications, but due to the mass availability of NLP tools, there are also opportunities for civilian black hats to use the technology in unsavory ways outside the scope of warfare. 
In the context of NLP, dual use has only recently been discussed in depth by~\citet{kaffee-etal-2023-thorny}. 
In their work, they define dual use in NLP as ``malicious reuse'' (\textit{i.e.}, \textbf{misuse}, where the intended purpose of the technology is violated). 
Examples of misuse across NLP include: manipulating models for automated influence operations (\textit{e.g.}, misinformation)~\cite{goldstein2023generative}, surveillance of marginalized groups \cite{sannon2022privacy}, and by-passing safety features to generate hate speech or otherwise engage in illegal activities (\textit{e.g.}, phishing) \cite{yong2024lowresource}. 
As black hat hackers are known to misuse white hat software~\cite{10.1145/3043955}, the threat of misuse against NLPSec research is palpable. 

To this end, \citet{kaffee-etal-2023-thorny}'s work focuses on traditional NLP, and thus the scope of their exploration is not configured to accommodate the 
unique position of NLPSec, as an interdisciplinary field. 
While the authors briefly mention \citet{10.1145/3600211.3604690} (who propose a defense method for preventing malicious use-cases of LLMs), as well as the possible criminal applications of LLMs for phishing, \citet{kaffee-etal-2023-thorny}'s proposed checklist does not help NLPSec practitioners to better navigate the problems of misuse.  
For example, their checklist for preventing misuse includes the following questions:
\begin{enumerate}[noitemsep,topsep=1pt]
    \item Can any scientific artifacts you create be used for military\footnote{Military funding has a long and complicated history in the sciences \cite{smit1995science}. While we do not examine the sources of funding across our sample of papers, we note the presence of institutions associated with the military and defense industry among the author affiliations.} application?
    \item  Can any scientific artifacts you create be used to harm or oppress any and particularly marginalised groups of society?
    \item Can any scientific artifacts you create be used to intentionally manipulate, such as spread disinformation or polarize people?
\end{enumerate}

In the context of NLPSec, the answer to the above questions will typically be ``yes''. 
Additionally, we observe that most papers only consider dual use or misuse as an afterthought, if at all, in line with the results of \citet{kaffee-etal-2023-thorny}.  
This concerning trend further
stresses the importance of a discussion on misuse, tailored to NLPSec.

\paragraph{Other Potential Harms in NLPSec}
The findings of our survey indicate that practitioners in NLPSec widely disagree about what potential harms are inherent to this research, making it difficult to ask NLPSec to blanketly minimize harm. 
Part of this nonagreement may stem from a historical understanding of how harm minimization is typically discussed in NLP.
Traditionally in NLP, 
practices of harm minimization have concerned very different issues than cybersecurity. Where cybersecurity is concerned with criminality, NLP has historically focused on fair payment of crowd-workers \cite{Shmueli2021BeyondFP}, and the prioritization of researching techniques that directly combat known flaws of LLMs \cite{weidinger2021ethical}, like 
dissemination of harmful social biases ~\cite{brown2020language,abid2021persistent,lucy-bamman-2021-gender} and misinformation~\cite{media_manipulation_20217, kenton2021alignment}. 
As discussed by \citet{leidner-plachouras-2017-ethical}, it is important that NLP researchers 
proactively plan to avoid unethical scenarios. 
As NLPSec combines the potential harms of NLP with those of cybersecurity, the need to anticipate and mitigate risks is crucial. 

To promote both effective and ethically responsible NLPSec research, we emphasize minimizing harm as a \textit{design principle}. Specifically, conversations about harm minimization should take place throughout a given project's research life-cycle, from the initial planning, funding, and designing of a project, to publishing and disseminating the conclusions \cite{Galinkin2022TowardsAR}.
This process ensures that research ethics remain core considerations throughout the work, rather than 
a mere rhetorical \textit{post-hoc} ethical statement \cite{9001063}.
Additionally, ~\citet{gardner2022ethical} emphasize the role of funding agencies in ensuring trustworthy AI, by \emph{mandating} ethical assessments throughout the application, evaluation and implementation phases, both from applicants and the funding agencies, aided by experts in ethics. 
Presently, such strict ethical requirements are not the norm in NLP, however. Until there are strong ethics review requirements across the field, as with other sciences, it is imperative that researchers clearly articulate the potential for dual use and misuse in their works, and provide viable defenses for the most vulnerable scenarios. 

In NLPSec, positive examples of harm minimization have prioritized user safety, avoiding scenarios which could expose sensitive data or scenarios that directly affect end-users. 
For example, ~\citet{parikh-etal-2022-canary}'s methodology intentionally avoids exposing sensitive data of real-world users in a data reconstruction attack by planting synthetic ``canaries'' (i.e., fake instances of private data) into their training data. 
Similarly, in their work exploring adversarial attacks, \citet{song-etal-2021-universal} do \textit{not} experiment with real-world systems, such that no end users are harmed.

\subsection{The Victims of English-Centric NLPSec}\label{subsec:multilinguality}

Below, we explore the role of multilinguality in NLPSec, 
the urgency of unresolved security vulnerabilities in relation to lower-resourced languages,  
and how traditional norms of consent in NLP may conflict with those of cybsersecurity.

\begin{figure}
    \centering
    \includegraphics[width=\linewidth]{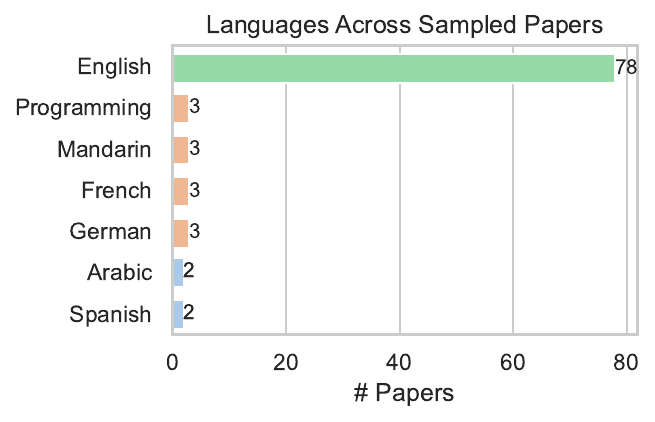}
    \caption{Of the victim languages investigated across the sample of 80 papers, we identify 22 \textit{natural} languages. 
    Languages not displayed above (as n=1) are Japanese \cite{zeng-xiong-2021-empirical}, Tibetan \cite{cao-etal-2023-pay-attention}, Javanese, Indonesian, Malaysian, Tagalog, Tamil \cite{wang2024backdoor}, and the remaining languages of the XNLI dataset: Greek, Bulgarian, Russian, Turkish, Vietnamese, Thai, Hindi, Swahili, and Urdu \cite{lin-etal-2024-inversion}.
    }
    \label{fig:languages}
\end{figure}

\paragraph{Security As Strong As Its Weakest Link}
As one of the most well represented languages in NLP, it is no surprise that English is present in 97\% of the works sampled (see Figure~\ref{fig:languages}). 
Emerging research examining multilingual NLPSec, however, suggests that that multilingual models may be \textit{more} vulnerable to attacks than their monolingual (English) counterparts, as demonstrated in the context of embedding inversion attacks~\cite{chen-y-etal-2024-text, chen-etal-2025-text} and backdoor attacks~\cite{he2024transferring}. 
Additionally, recent works also show how lower-resourced languages can be weaponized to bypass LLM safety features~\cite{yong2024lowresource}, as well as introduce backdoors~\cite{wang2024backdoor}, creating further cause for concern. To this end, \citet{yong2024lowresource} discuss the apparent shift in consequences for poor performance over lower-resourced languages: previously, a lack of competitive models to handle these languages culminated primarily in technological disparity, affecting only the community in question. This inequality can be exploited by malicious actors, resulting in a threat to everyone. In other words, the security of NLP models is now only as strong as its weakest link. Alarmingly, such weaponization of under-performing language technologies is already being observed \cite{Nigatu_2024}.

\paragraph{Higher Stakes For Lower Resourced Scenarios}
Low-resource languages can often be situated in vulnerable contexts, which brings heightened need to protect the communities speaking them.
For example, previous works in NLP have exposed correlations between GDP and data availability \cite{blasi-etal-2022-systematic, ranathunga-de-silva-2022-languages}, underscoring how the gap between higher- and lower-resource languages is part of a broader picture of global inequality. 
At the same time, even within wealthier nations, minority languages (often low-resource) may require revitalization efforts in order to stave off extinction (\textit{e.g.}, Indigenous languages of Australia ~\citep{meakins2016loss}); and other widely-adopted languages may battle stigma, obstructing their inclusion in language technology (\textit{e.g.}, Creoles ~\citep{lent2024creoleval}). 
Practitioners in low-resource NLP have developed their own ethical norms in response to such concerns, grounded in the prioritization of a community's specific needs and aspirations for language technology, as well as the preservation of their autonomy \cite{bird-2022-local, lent-etal-2022-creole, mager-etal-2023-ethical}.

\paragraph{Research Traditions Collide: Consent} 

Consent is highly context-dependent in both cybersecurity and general NLP research ethics, leading to potential conflicts when these fields intersect in NLPSec. 
In traditional cybersecurity, the necessity of consent largely depends on the nature of the system being tested. 
When testing systems running on third-party infrastructure, such as web services or cloud platforms, obtaining explicit consent is expected in order to, \textit{e.g.}, avoid legal or ethical issues. 
However, consent is not typically required for, \textit{e.g.}, research involving hardware or software running locally on the researcher’s infrastructure, even if it violates end-user license agreements (EULAs) or terms-of-service (ToS) contracts \cite{kozhuharova2022ethics}. 
For example, reverse engineering or probing locally deployed systems for vulnerabilities is widely accepted as a valid and necessary practice, provided it prioritizes public safety and minimizes harm.

In contrast, multilingual NLP research has increasingly emphasized \textit{community} consent, particularly regarding low-resource languages or marginalized groups. 
These efforts are grounded in the principle of respecting the autonomy and cultural context of the communities whose languages and data are being studied. 
NLPSec introduces scenarios where these norms may conflict. 
For example, securing LLMs for low-resource languages is vital for ensuring downstream safety of their communities. 
However, using a language for security testing without explicit consent risks reducing it to a mere tool for experimentation, potentially alienating the communities involved and exacerbating existing inequalities \citep{bird-2020-decolonising}.

This aspect is largely ignored in NLPSec to date. 
Among our sample, only one work engaged with a truly low-resource language (see Figure~\ref{fig:languages}). 
In this study, \citet{cao-etal-2023-pay-attention} aim to raise awareness about the threatened security of minority languages by contributing a script-based adversarial attack method for Tibetan, a notably vulnerable language\footnote{Tibetan has relatively few speakers (1.2 million native speakers according to \url{https://en.wikipedia.org/wiki/Lhasa_Tibetan} and 6 million speakers in total~\cite{Tournadre2013TheTL}), is actively undergoing revitalization \cite{roche2018language}, and is situated in a tender sociopolitical context~\cite{Roche2017IntroductionTT, Jia2021ASA}.}  against CINO~\cite{yang-etal-2022-cino}\footnote{CINO a pre-trained multilingual LLM for handling languages spoken across China (\textit{i.e.}, Mandarin, Cantonese, Korean, Mongolian, Uyghur, Kazakh, Zhuang, and Tibetan).}.

Ultimately, the ``white hat'' versus ``black hat'' paradigm in cybersecurity is further complicated when extended to NLPSec. While cybersecurity often frames consent as an ethical trade-off, NLPSec researchers must also mind the long-standing ethical norms of NLP, particularly for lower-resourced or otherwise marginalized languages. 
Balancing these differing research norms around consent demands careful consideration. 
On the one hand, engaging with language community representatives and aligning research with their needs can ensure that low-resource languages are not exploited in ways that harm or undermine their speakers. 
On the other hand, delaying research to secure consent could leave vulnerable languages at greater risk of exploitation by malicious actors.
Given the present and severe threat of weaponization of lower-resourced languages, however, this issue must not remain unaddressed by NLPSec; prioritization of ethical research practices will be critical for harm minimization.

\subsection{Obstacles for CVD in NLPSec}\label{sec:cvd-revisited}
The results of our survey showed that no works \textit{report} whether efforts to do CVD (Coordinated Vulnerability Disclosure, see Section~\ref{sec:cvd}) occurred. 
Outside the scope of our survey, positive examples of CVD in NLPSec do exist, such as \citet{carlini2024stealing}, who state clearly in their manuscript that the discovered vulnerability was reported to OpenAI and that the release of the paper followed the company's response to and mitigation of the risk described in their work. 
In this section, we aim to explore some reasons why CVD is a potentially nontrivial in NLPSec.

\paragraph{Not Fixable by One Line of Code}
Models are one part of complex computer software systems subject to more general cybersecurity vulnerabilities, \textit{e.g.}, remote code execution, as has been observed in the \texttt{llama-cpp-python} library.\footnote{\url{https://github.com/abetlen/llama-cpp-python/security/advisories/GHSA-56xg-wfcc-g829}}
In such instances, CVD offers NLPSec researchers a way to maintain transparency, without assisting malicious actors, as the published vulnerabilities will hopefully have already been patched. 
In NLP and the broader machine learning space, however, this remediation and mitigation process is complicated by the fact that \textbf{discovered issues may be endemic} to the target of evaluation or require prohibitively expensive retraining to fix.
As opposed to traditional software, one cannot simply write a patch that fixes a discovered issue entirely, and instead, guidance must be provided to users in order to allow them to accept or mitigate risk appropriately.
In one recent example, the LAION 5B dataset was found to contain child sex abuse material~\cite{birhane2021multimodaldatasetsmisogynypornography, birhane2023laionsdeninvestigatinghate}. 
The dataset was accordingly taken offline by its authors \cite{laion2023safety, thiel2023laoin}.
Consequently, all models trained on this dataset were at known risk of producing illegal, harmful materials.
Model providers, upon being made aware of this risk, had the obligation to decide whether to retrain the model or accept the risk -- this was not something that could be managed by updating a few lines of code. 
Still, disclosure of the risk allows affected parties to make informed decisions.

\paragraph{An Open Problem for Open Models}
While CVD is most relevant for research using proprietary models, it also remains relevant for open-weight, freely available models. 
As the majority of works in NLPSec have thus far been concerned with attacking or defending open-weight models (see Figure~\ref{fig:victim}), CVD may seem less applicable to models that are not actively maintained by the organization hosting them. 
In such a case where the practical steps towards CVD may be unclear, an open question in NLPSec is how to best disclose risks
-- if at all -- 
both to the pertinent organizations and to the broader scope of users. 
Similar to the trolley problem introduced by \citet{Kohno-2023} (Section~\ref{sec:cvd}), there may be instances where CVD requires careful consideration, for example in critical sectors like healthcare.  
However in the majority of NLP, where a model can typically be replaced with another with relative ease, it is difficult to give a generalized conjecture on such cases.   
For our part, we recommend that NLP researchers engage in best-effort attempts to alert model providers to potential risks. 
Most companies that provide models as a service have a channel for external users to file bug reports.
In cases where the model is produced by a smaller entity, opening issues on the platform where the model is shared \textit{e.g.} Github, HuggingFace or emailing authors of a paper tied to the model serves as a good channel for attempting this coordination before publication of results.

Another complication for CVD in NLPSec is scalability. As a field, NLP places immense value on scalability \cite{kogkalidis2024tablesnumbersnumbers}. Researchers are often expected or encouraged to massively scale their experiments to an increasing number of models and languages. While scaling experiments largely hinges upon the availability of compute resources, the process of CVD does not scale so. Ideally, CVD entails intentional, personal communication between the researcher and the affected organization. Responsible disclosure to CERTs or other trusted communities functions the same way. The human aspect of this process cannot simply be outsourced and automated to a machine, thus conflicting with the expectations for massive scalability within NLPSec. 

\subsection{Recommendations for NLPSec Practitioners}\label{sec:recommendations}

Thusfar, the field of NLPSec has largely been operating in a ``gray hat'' manner, where the individual researcher is compelled to rely on their own moral compass. This is in part due to the overwhelming bulk of AI regulatory documents \cite{larsson2021ai}, which often do not directly relate to a security angle, as well as the rigidity of applying certain cyber security ethical norms to the unique problems of NLPSec. 
In response to this pressure point, we aim to provide some
concrete recommendations to help the field take concrete steps towards more ethical NLPSec. 
To this end, we hope future works will benefit from, and build upon, the following recommendations:

\begin{enumerate}
    
    \item \textbf{Plan Ahead to Minimize Harm:}
    Ethical considerations should not be relegated to a post-hoc ethics statement. First, consider the harms entailed in conducting a study and in foregoing it. Design experiments with harm minimization in mind from the start. Include these details in the main body of your work. Beyond reducing the potential harms of research and helping researchers avoid downstream ethical conundrums, this approach also promotes a culture of responsible research.   

    \item \textbf{Prioritize Multilingual Equity:}
    Include multilingual models and lower-resourced languages in NLPSec work to build towards comprehensive security coverage for all. Prioritize typologically diverse language samples \cite{ploeger-etal-2024-framework}. Engage with the communities speaking these languages to seek consent and avoid exploitation. Consider whether a particular community might be jeopardized as a result of your work. Researchers should respect the autonomy of these communities, while working to address the established heightened vulnerabilities in such low-resource scenarios.   

    \item \textbf{Approach Disclosure Responsibly:} 
    
    \textbf{(a)} Consider the most appropriate options for disclosure. If you can complete CVD, contact relevant parties about security breaches 60-90 days prior to any publication and clearly acknowledge that CVD occurred directly in the published manuscript. Even for open-weight models, best-effort attempts to alert stakeholders should be made, such as model providers or the platforms hosting the models. If you cannot complete CVD, attempt responsible disclosure to other affected parties. Decide whether public disclosure is appropriate, and act accordingly. Communicate this thought process in your paper.
    \textbf{(b)} When appropriate, release accompanying proof-of-concept code to help NLPSec researchers better defend against attacks. While black hats will always make time to re-implement attacks for nefarious gains, white hats are time-constrained and defense becomes harder without clear \& explicit resources. If not appropriate, explain why in your manuscript. Ask yourself if public disclosure is still warranted, if open-sourcing code is not.
\end{enumerate}

\section{Conclusion}
In the burgeoning field of NLPSec, most works consider scenarios where a malicious attacker seeks to undermine a system's intended behavior, with the goal of causing harm.
Research output in NLPSec thus stands to be highly consequential in the face of mass-adoption of language technologies such as LLMs, and its relevance to public safety necessitates heightened scrutiny when it comes to best practices for ethical research. 
Given NLPSec's position as a truly interdisciplinary field, practitioners in this space can benefit from the rich traditions of research ethics from both cybersecurity and NLP. 
In this work, however, we find that NLPSec works published in NLP venues generally fall short of the ethical standards set by cybersecurity (Section~\ref{sec:nlp}), signaling a higher-level disconnect between NLP and cybersecurity practitioners for work in this area.
This failure to inherit ethical best practices can arise from a variety factors (Section~\ref{sec:discussion}), but largely stem from the differences between traditional cybersecurity and NLPSec, which underscores the 
limitations of the ``white versus black hat'' paradigm of cybersecurity as applied to NLPSec. 
Still, we argue that the repercussions of the current research patterns are grim: works in NLPSec may benefit would-be attackers more than the public (Section~\ref{sec:cybersec}), with dire consequences for everyone, but especially for already-marginalized communities (Section~\ref{subsec:multilinguality}). 
By highlighting these problems and exploring their nuances, this work aims to 
persuade the field of the urgency of the present situation and to
spark a much-needed conversation across the field of NLPSec. 
To kick off this conversation, we provide some concrete recommendations to help practitioners transition from gray hat to white hat NLP. 

\section*{Limitations}

\paragraph{Defining Ethical Hacking}
While ethical hacking is, on its face, a noble venture, it is \textbf{a term that features some subjectivity and may find itself at odds with the desires of particular groups or individuals}. 
For instance, the definition of an ethical hacker as one who is ``trustworthy for business and lawful''~\cite{christen2020ethics} may run headlong into both trust and the law.
Regarding trust, many organizations have adopted an approach that is far more friendly to security researchers, but there are organizations who are notorious to this day for their attempts to keep the discovery of vulnerabilities in their products quiet. 
As it concerns the law, there are two primary issues to contend with.
First, what is ``lawful'' will necessarily change across jurisdictions, with laws differing not merely between countries, but sometimes across provinces and states.
For example, the ethics of government-associated cybersecurity research (\textit{e.g.} government hacking) can be a topic of debate, even when practitioners are acting under the color of law.
Another example includes the
exemption for security research in the United States, which is not written into law, but is rather part of the US Department of Justice's prosecution guidelines~\cite{doj-2022-cfaa}, updated in 2022, indicating that good faith security research should not be prosecuted. 
In other words, much ethical hacking in the US may be considered unlawful but will simply not be prosecuted.
The second issue is time. Across jurisdictions, laws are likewise positioned to evolve over time, especially as the list of known cyber-threats grows to include attacks against NLP models. 
In general, as it is often difficult for the law to keep up with rapid technological progress, ethics training must be prioritized in both the NLP \cite{bender-etal-2020-integrating} and cybersecurity curriculum \cite{BlankenWebb2018ACS}. 

\paragraph{Risks of Monocultural Ethics Discourse}
As AI becomes further entrenched in daily life, the risks imposed from research and commercial activities in AI are also a global issue. 
Similar to how correspondents of~\citet{kaffee-etal-2023-thorny}'s survey are overwhelmingly from a western audience\footnote{Of 48 participants, only 3 hailed from Asia and 1 from Africa, with the remainder from Europe or North America.}, the bulk of AI governance documents are also overwhelmingly of Western origin (e.g.~\citealp{larsson2021ai, unesco2021recommendation}).
In contrast, Figure~\ref{fig:authors} (Appendix \ref{appendix:details}) reveals that the majority of NLPSec research comes from Asia.
However, 
historical and cultural differences have led to fundamentally different approaches to addressing AI risks across these regions.\footnote{The authors acknowledge that we represent Western institutions and have our own values and biases accordingly.}
For example, while the deployment of technologies such as facial recognition is illegal and considered strictly unethical in the EU because of GDPR~\citep{EuropeanParliament2016a}, it is widely deployed in countries such as China~\citep{chinafacial}, Iran~\citep{iran_ai_woman_2023}, 
Canada \cite{canadafacial}, 
and the US \cite{usafacial}, where the local personal data is collected, raising concerns over human rights. 
Still, models trained on such data may be imported to the EU and deployed without any legal consequences, highlighting a global ethical risk, termed \textit{ethics dumping}~\citep{ec2013horizon,european2016h2020}, where non-ethical practices are shifted to countries lacking certain ethics regulations.
This divide between AI governance and the regions impacted by AI calls for inclusion of diverse perspectives, especially in a burgeoning and cross-disciplinary field like NLPSec.
Of course, cross-cultural AI ethics is notably diverse. For example, African Ubuntu philosophy promotes communal values in the use of AI~\citep{gwagwa2022role}, Abrahamic religious views stress that AI use should respect human dignity \cite{10.1093/ojlr/rwaa015, Raquib2022IslamicVE}, and Buddhist AI philosophy advocates for reducing pain and suffering using AI \cite{hughes2012compassionate, buddhism_ai}. 
When faced with seemingly irresolvable conflicts of ethical values across cultures, we urge NLPSec researchers to look towards the UN Declaration of Human Rights, 
which outlines the fundamental rights and freedoms of all human beings.

\section*{Ethics Statement}
Our work adheres to the ACM Code of Ethics.
As we analyze, \textit{e.g.} author metadata, we have ensured that the licenses of the data sources allow for this type of data extraction.
This line of work, including our methodology and the analysis of author meta-data, has received approval from the  Aalborg University Research Ethics Committee under case number 2024-505-00376.

\section*{Acknowledgments}
HL, YC, and JB are funded by the Carlsberg Foundation, under the Semper Ardens: Accelerate programme (project nr. CF21-0454). 
This work benefited greatly from help and conversations with others. 
Thank you to Christopher Fiorelli and Maria Antoniak at AI2 for their help with the Semantic Scholar crawler,
to Zeerak Talat for detailed feedback on our manuscript,
and to Steven Bird for conversations on ethics of AI, which enriched this manuscript. 
Thank you to members of the AAU NLP Research Group for their feedback during paper clinics,
to Mike Zhang for help improving figures, to Nicholas Walker for feedback on multiple drafts of this paper,
and to Shreyas Srinivasa for initial input on cybersecurity research norms. 
Finally, thank you to the TACL Area Chair and Reviewers, whose constructive feedback played a major role in shaping this manuscript.

\bibliography{anthology,custom}
\bibliographystyle{acl_natbib}

\newpage

\appendix

\section{Reproducibility of Results}\label{appendix:papers}
We provide our exact annotations for transparency and reproducibility.
The papers sampled in this work are listed in Tables~\ref{tab:adversarial} (adversarial attacks), Table~\ref{tab:backdoors} (backdoor attacks), and Table~\ref{tab:data reconstruction} (Data Reconstruction attacks, which include both embedding inversion attacks and instance encoding/embedding encryption), below. Among these 80 papers, we found zero discussion of dual use and zero instances of coordinated vulnerability disclosure (CVD), so we do not list out the annotations for these two categories. 

\begin{table*}[]
\small
\centering

  \resizebox{\textwidth}{!}{  
\begin{tabular}{llllll}
\toprule
\textbf{Paper} & \textbf{Contribution} &  \textbf{Ethics} & \textbf{Misuse} & \textbf{Open Source} & \textbf{Languages} \\  \midrule
\citet{ren-etal-2019-generating} & Attack & No & No & Yes & Eng \\
\citet{wallace-etal-2019-universal} & Attack & No & No & Yes & Eng \\
\citet{han-etal-2020-adversarial} & Both & No & No & Yes & Eng \\
\citet{zang-etal-2020-word} & Attack & No & No & Yes & Eng \\
\citet{li-etal-2020-bert-attack} & Attack & No & No & Yes & Eng \\
\citet{xu-etal-2021-grey} & Both & Yes & Yes & Yes & Eng \\
\citet{zeng-etal-2021-openattack} & Attack & Yes & Yes & Yes & Eng, Zho \\
\citet{chen-etal-2021-multi} & Attack & Yes & Yes & Yes & Eng \\
\citet{song-etal-2021-universal} & Attack & Yes & Yes & Yes & Eng \\
\citet{li-etal-2021-contextualized} & Attack & No & No & Yes & Eng \\
\citet{zhou-etal-2021-defense} & Defense & No & No & No & Eng \\
\citet{keller-etal-2021-bert} & Defense & No & No & Yes & Eng \\
\citet{zeng-xiong-2021-empirical} & Attack & No & No & No & Eng, Zho, Jpn \\
\citet{swenor-kalita-2021-using} & Defense & No & No & No & Eng \\
\citet{bao-etal-2021-defending} & Defense & No & No & Yes & Eng \\
\citet{raina-gales-2022-residue} & Defense & Yes & No & Yes & Eng \\
\citet{choi-etal-2022-tabs} & Attack & No & No & No & Eng, Prog. \\
\citet{lei-etal-2022-phrase} & Attack & No & No & Yes & Eng \\
\citet{xu-etal-2022-weight} & Defense & No & No & No & Eng \\
\citet{xie-etal-2022-word} & Attack & Yes & No & Yes & Eng \\
\citet{fang-etal-2023-modeling} & Attack & Yes & No & Yes & Eng \\
\citet{cao-etal-2023-pay-attention} & Attack & Yes & No & Yes & Bod \\
\citet{li-etal-2023-text} & Defense & No & No & No & Eng \\
\citet{wang-etal-2023-rmlm} & Defense & No & No & No & Eng \\
\citet{tsymboi-etal-2023-layerwise} & Attack & No & Yes & Yes & Eng \\
\citet{gao-etal-2024-semantic} & Attack & No & No & No & Eng \\
\citet{sadrizadeh-etal-2024-classification} & Attack & Yes & Yes & Yes & Eng, Fra, Deu \\
\citet{zhang-etal-2024-random} & Defense & No & No & No & Eng \\
\citet{chen-etal-2024-context} & Attack & No & No & No & Eng \\
\citet{wang-etal-2024-da3} & Attack & Yes & Yes & Yes & Eng \\
\citet{xu-wang-2024-linkprompt} & Attack & Yes & Yes & Yes & Eng \\
\citet{yu-etal-2024-query} & Attack & No & No & Yes & Eng \\
\citet{alshahrani-etal-2024-arabic} & Attack & No & No & Yes & Ara \\
\bottomrule
\end{tabular}
}
\caption{Papers sampled pertaining to \textbf{adversarial attacks}. Under languages, ``Prog.'' is short for programming language(s).}
\label{tab:adversarial}
\end{table*}

\begin{table*}[]
\small
\centering
  \resizebox{\textwidth}{!}{  
\begin{tabular}{llllll}
\toprule
\textbf{Paper} & \textbf{Contribution} & \textbf{Ethics} & \textbf{Misuse} & \textbf{Open Source} & \textbf{Languages} \\  \midrule
\citet{yang-etal-2021-rethinking} & Attack & Yes & No & Yes & Eng \\
\citet{qi-etal-2021-hidden} & Attack & Yes & Yes & Yes & Eng \\
\citet{qi-etal-2021-mind} & Attack & Yes & Yes & Yes & Eng \\
\citet{qi-etal-2021-turn} & Attack & Yes & Yes & Yes & Eng \\
\citet{yang-etal-2021-rap} & Defense & Yes & Yes & Yes & Eng \\
\citet{qi-etal-2021-onion} & Defense & Yes & Yes & Yes & Eng \\
\cite{li-etal-2021-bfclass-backdoor} & Defense & Yes & Yes & Empty repo & Eng \\
\citet{li-etal-2021-backdoor} & Attack & No & No & No & Eng \\
\citet{chen-etal-2022-textual} & Attack & Yes & Yes & Yes & Eng \\
\citet{yoo-kwak-2022-backdoor} & Attack & No & No & No & Eng \\
\citet{gan-etal-2022-triggerless} & Attack & Yes & Yes & Yes & Eng \\
\citet{chen-etal-2022-expose} & Defense & Yes & No & Yes & Eng \\
\citet{zhang-etal-2022-fine-mixing} & Defense & No & No & No & Eng \\
\citet{lyu-etal-2022-study} & Defense & No & No & Yes & Eng \\
\citet{jin-etal-2022-wedef} & Defense & Yes & Yes & Broken link & Eng \\
\citet{zhang-etal-2022-dim} & Defense & Yes & No & No & Eng \\
\citet{liu-etal-2023-maximum} & Defense & No & No & No & Eng \\
\citet{zhao-etal-2023-prompt} & Attack & Yes & Yes & Yes & Eng \\
\citet{mei-etal-2023-notable} & Attack & Yes & Yes & Yes & Eng \\
\citet{he-etal-2023-imbert} & Defense & No & No & Yes & Eng \\
\citet{you-etal-2023-large} & Both & No & No & No & Eng \\
\citet{li-etal-2023-multi-target} & Attack & Yes & Yes & Yes & Eng, Prog. \\
\citet{yan-etal-2023-bite} & Both & Yes & Yes & Yes & Eng \\
\citet{li-etal-2023-defending} & Defense & Yes & Yes & Yes & Eng \\
\citet{he-etal-2023-mitigating} & Defense & No & No & Yes & Eng \\
\citet{huang-etal-2024-composite} & Attack & Yes & Yes & Yes & Eng \\
\citet{li-etal-2024-chatgpt} & Attack & Yes & Yes & Yes & Eng \\
\citet{du-etal-2024-uor} & Attack & Yes & Yes & No & Eng \\
\citet{graf-etal-2024-two} & Defense & Yes & Yes & Empty repo & Eng \\
\citet{zeng-etal-2024-beear} & Defense & Yes & Yes & Broken link & Eng, Prog. \\
\citet{yi-etal-2024-badacts} & Defense & Yes & No & Yes & Eng \\
\citet{wu-etal-2024-acquiring} & Defense & Yes & No & Yes & Eng \\
\citet{wang-etal-2024-backdoor} & Attack & No & No & No & \begin{tabular}[c]{@{}l@{}}Eng, Jav, Ind, \\ Msa, Tgl, Tam \\\end{tabular} \\
\bottomrule
\end{tabular}
}
\caption{Papers sampled pertaining to \textbf{backdoor attacks}. Under languages, ``Prog.'' is short for programming languages.}
\label{tab:backdoors}
\end{table*}

\begin{table*}[]
\small
\centering
  \resizebox{\textwidth}{!}{  
\begin{tabular}{llllll}
\toprule
\textbf{Paper} & \textbf{Contribution} &  \textbf{Ethics} & \textbf{Misuse} & \textbf{Open Source} & \textbf{Languages} \\ \midrule
\citet{huang-etal-2020-texthide} & Defense & No & No & Yes & Eng \\
\citet{xie-hong-2021-reconstruction} & Attack & Yes & Yes & No & Eng \\
\citet{xie-hong-2022-differentially} & Defense & No & No & No & Eng \\
\citet{hayet-etal-2022-invernet} & Attack & No & No & Yes & Eng \\
\citet{parikh-etal-2022-canary} & Both & Yes & Yes & No & Eng \\
\citet{kim-etal-2022-toward} & Defense & No & No & No & Eng \\
\citet{morris-etal-2023-text} & Attack & No & No & Yes & Eng \\
\citet{li-etal-2023-sentence} & Attack & Yes & No & Yes & Eng \\
\citet{zhou-etal-2023-textobfuscator} & Defense & No & No & Yes & Eng \\
\citet{zhang-etal-2023-fedpetuning} & Defense & No & No & Yes & Eng \\
\citet{chen-y-etal-2024-text} & Both & Yes & Yes & Yes & Eng, Fra, Deu, Spa \\
\citet{huang-etal-2024-transferable} & Attack & No & No & Empty repo & Eng \\
\citet{elmahdy-salem-2024-deconstructing} & Attack & No & No & No & Eng \\
\citet{lin-etal-2024-inversion} & Attack & No & No & No & \begin{tabular}[c]{@{}l@{}}Eng, Fra, Deu, \\ Spa, Ell, Bul, \\ Rus, Tur, Ara, \\ Vie, Tha, Zho, \\ Hin, Swa, Urd \\ \end{tabular} \\
\bottomrule
\end{tabular}
}
\caption{Papers sampled pertaining to \textbf{data reconstruction attacks}, which includes both \textbf{embedding inversion} and \textbf{embedding encryption} (i.e., instance encoding).}
\label{tab:data reconstruction}
\end{table*}

\section{Complementary Results}\label{appendix:details}
\begin{figure}[h]
    \centering
    \includegraphics[width = 0.8\linewidth]{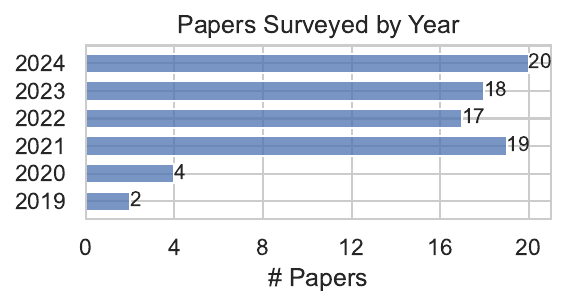}
    \caption{Distribution of sampled papers across their year of publication.}
    \label{fig:year}
\end{figure}

\begin{figure}[h]
    \centering
    \includegraphics[width = \linewidth]{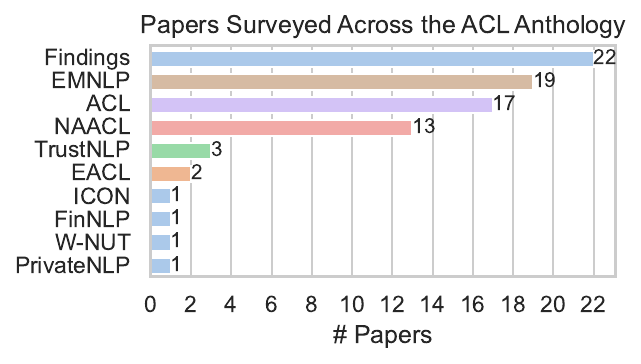}
    \caption{Distribution of sampled papers published in different venues across the ACL* community. For the Findings papers specifically, 11 are at ACL, 8 at EMNLP, 2 at NAACL, and 1 at EACL.}
    \label{fig:venue}
\end{figure}

\begin{figure}[h]
    \centering
    \includegraphics[width = \linewidth]{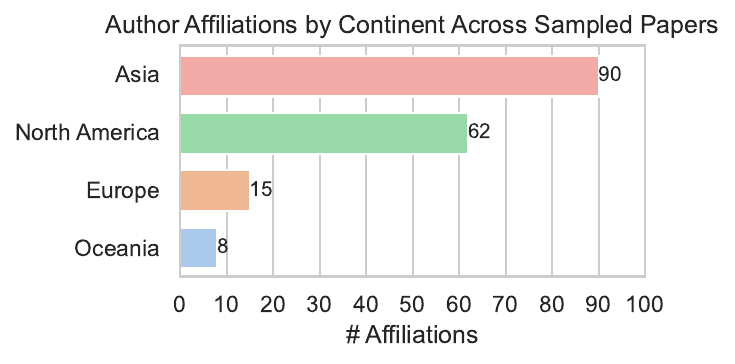}
    \caption{We show the distribution of author affiliations across continents, rather than countries, for easy comparison against ~\citet{kaffee-etal-2023-thorny}). Most research from our sample of NLPSec works hails from institutions residing in Asia. Conversely, governance documents come overwhelmingly from the EU and US~\cite{Jobin2019TheGL}, and discussions on dual-use and ethics in NLP are also Western-centric (n=3 from Asia and n=1 from Africa in a survey of 48 people by~\citet{kaffee-etal-2023-thorny}). This highlights the need for a wider dialogue across the field. When cross-cultural values conflict and best practices become unclear, practitioners should consider the UN Declaration of Human Rights.
    }
    \label{fig:authors}
\end{figure}

\begin{figure}[h]
    \centering
    \includegraphics[width=\linewidth]{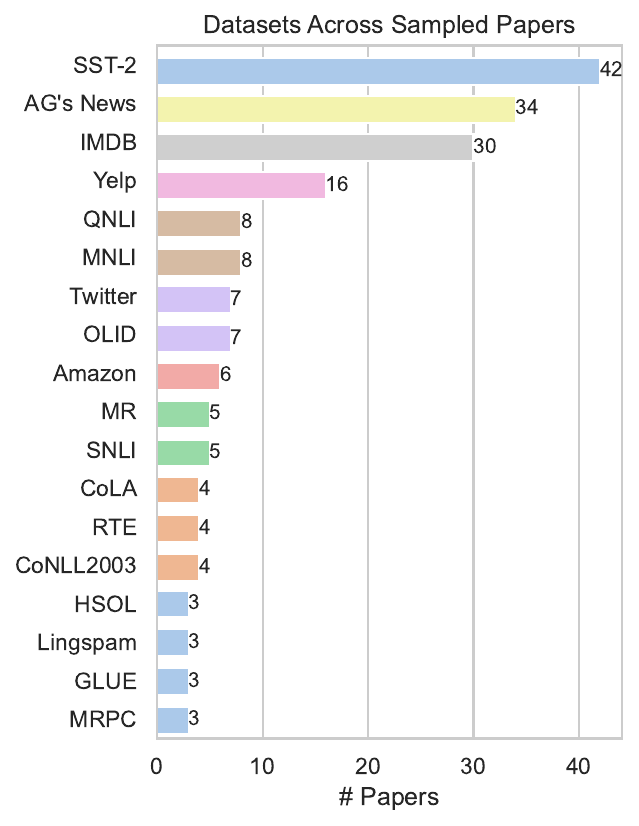}
    \caption{Distribution of datasets used to validate experiments across the sampled works in NLPSec. For the purposes of visualization, we do not display the long tail of datasets which were used by only one paper.
    }
    \label{fig:datasets}
\end{figure}

In this Appendix, we provide more general information about the sampled papers, for further transparency. 
First, Figure~\ref{fig:year} shows the relative age of publications, and Figure~\ref{fig:venue} shows their distribution across venues within the ACL* community. 
Figure~\ref{fig:authors} presents the continent associated with every unique affiliation in the author list, so that we may provide rough demographic information about our sample. Finally, Figure~\ref{fig:datasets} show the most common evaluation datasets across our sample of works in NLPSec.

\end{document}